\documentclass{article}
\usepackage{arxiv}
\usepackage[utf8]{inputenc} 
\usepackage[T1]{fontenc}    
\usepackage{hyperref}       
\usepackage{url}            
\usepackage{booktabs}       
\usepackage{amsfonts}       
\usepackage{nicefrac}       
\usepackage{microtype}      
\usepackage{lipsum}		
\usepackage{graphicx}
\usepackage{natbib}
\usepackage{doi}
\usepackage{amsmath} 
\usepackage{breqn}
\usepackage{flexisym}
\usepackage{mathtools}
\usepackage{caption}
\usepackage{graphicx}
\usepackage{float} 

\usepackage{subfig}
\hypersetup{hypertex=true,
    colorlinks=true,
    linkcolor=ccr,
    anchorcolor=ccr,
    citecolor=ccr}

\title{GUNet: A Graph Convolutional Network United Diffusion Model for Stable and Diversity Pose Generation}
\date{} 					
\date{}

\usepackage{authblk}

\author[1]{Shuowen Liang\textsuperscript{*}}
\author[2]{Sisi Li\textsuperscript{*}}
\author[2]{Qingyun Wang\textsuperscript{†}}
\author[2]{Cen Zhang}
\author[2]{Kaiquan Zhu}
\author[2]{, Tian Yang}
\affil[1]{School of Electronic Information Engineering, Beijing Jiaotong University, China,22120077@bjtu.edu.cn}
\affil[2]{Sohu.com,{\{sisili,qingyunwang217513,cenzhang117649,kaiquanzhu,tianyang\}@sohu-inc.com}}

\renewcommand{\headeright}{ }
\renewcommand{\undertitle}{ }
\renewcommand{\shorttitle}{GUNet: A Graph Convolutional Network United Diffusion Model for Stable and Diversity Pose Generation}

\begin{document}
\footnotetext[1]{\textsuperscript{*}Equal contribution}
\footnotetext[2]{\textsuperscript{†}Corresponding author}
\definecolor{ccr}{RGB}{10,110,150}
\maketitle
\begin{abstract}
Pose skeleton images are an important reference in pose-controllable image generation. In order to enrich the source of skeleton images, recent works have investigated the generation of pose skeletons based on natural language. These methods are based on GANs. However, it remains challenging to perform diverse, structurally correct and aesthetically pleasing human pose skeleton generation with various textual inputs. To address this problem, we propose a framework with \textbf{GUNet} as the main model, \textbf{PoseDiffusion}. It is the first generative framework based on a diffusion model and also contains a series of variants fine-tuned based on a stable diffusion model. PoseDiffusion demonstrates several desired properties that outperform existing methods. \textit{1) Correct Skeletons.} GUNet, a denoising model of PoseDiffusion, is designed to incorporate graphical convolutional neural networks. It is able to learn the spatial relationships of the human skeleton by introducing skeletal information during the training process. \textit{2) Diversity.} We decouple the key points of the skeleton and characterise them separately, and use cross-attention to introduce textual conditions. Experimental results show that PoseDiffusion outperforms existing SoTA algorithms in terms of stability and diversity of text-driven pose skeleton generation. Qualitative analyses further demonstrate its superiority for controllable generation in Stable Diffusion. Homepage:\url{https://github.com/LIANGSHUOWEN/PoseDiffusion}
\end{abstract}


\section{Introduction}

As an important external control condition in controllable image generation, 2D pose skeleton images are critical to the quality of the generated images. For example, ControlNet(\cite{zhang2023adding}), HumanSD(\cite{ju2023humansd}), GRPose(\cite{yin2024grpose}) and other models used to create controllable photos need to be provided with one or more 2D pose skeleton images for reference. In addition, many pose sequence generation tasks also need to be provided with the first frame of the pose skeleton. However, in order to obtain the pose skeletons, current methods rely mainly on extracting them from existing images using pose detection models, such as DWPose(\cite{yang2023effective}) and OpenPose(\cite{cao2017realtime}). This limits the diversity and operability of obtainable pose skeletons. In practice, there are also blender-based applications which allow users to adjust and model the pose by dragging and dropping key points. However, these approaches are time-consuming and require a lot of manual effort. They are also difficult to implement an end-to-end process due to the large amount of manual intervention. In order to get pose skeleton images more easily and flexibly, it is necessary to create a framework for generating pose skeletons directly from natural language. 

\begin{figure}[tbp]
	\centering
	\subfloat[]{\label{fig:1(a)}\includegraphics[width=0.17\textwidth]{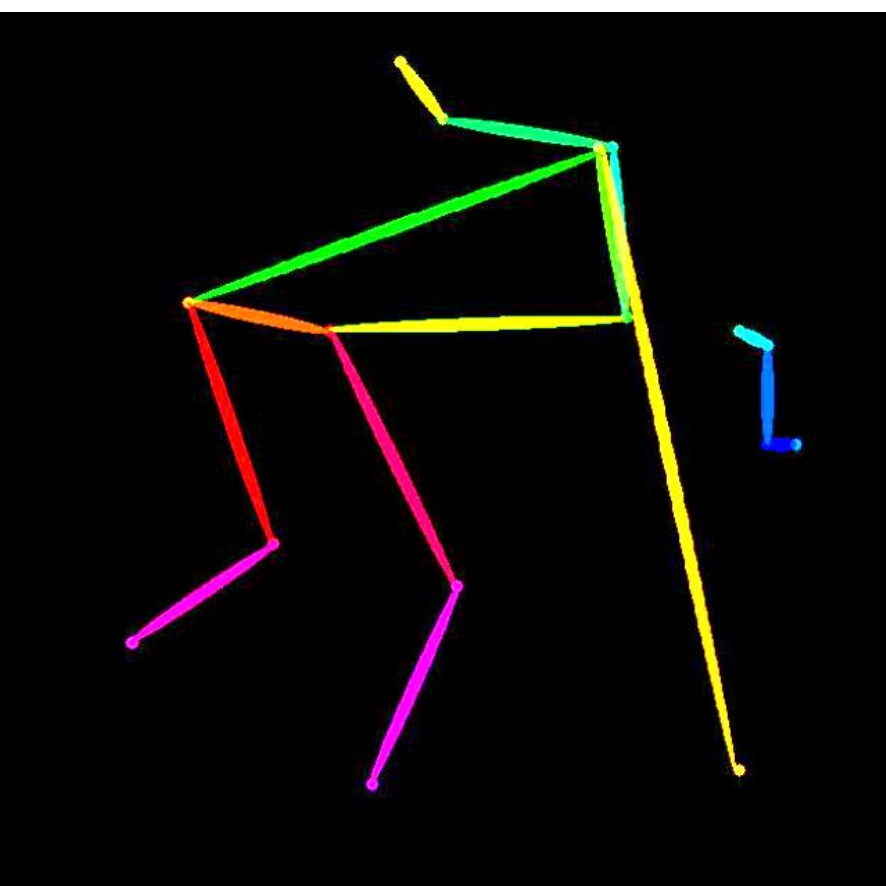}}\quad
	\subfloat[]{\label{fig:1(b)}\includegraphics[width=0.17\textwidth]{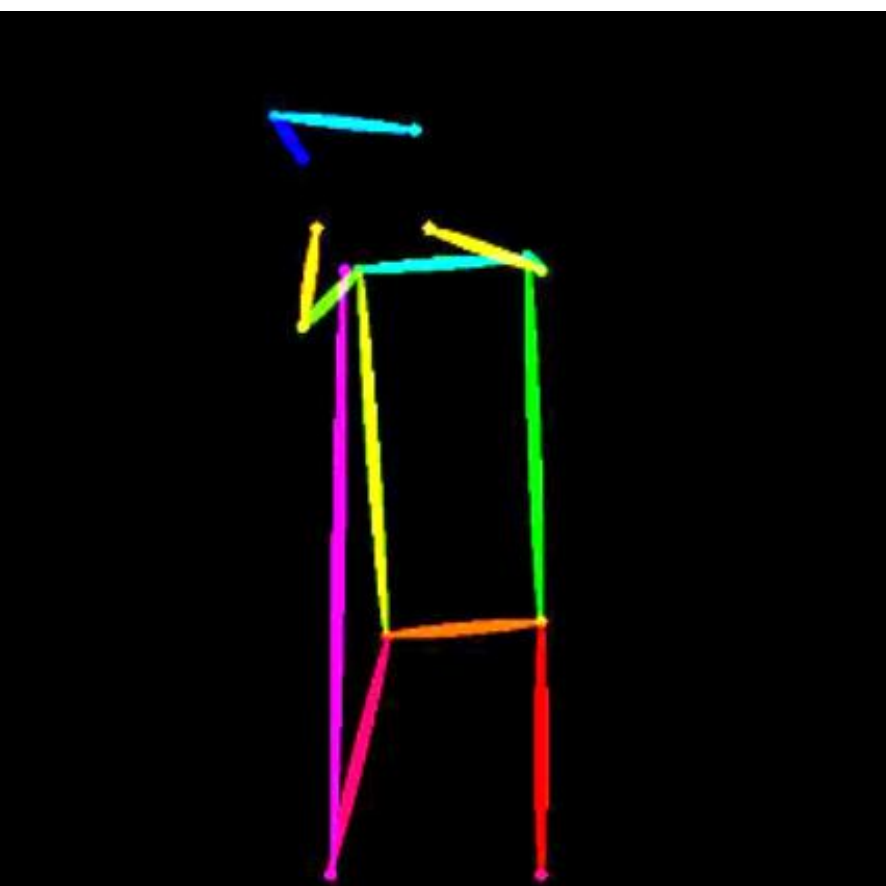}}\quad	
        \subfloat[]{\label{fig:1(c)}\includegraphics[width=0.17\textwidth]{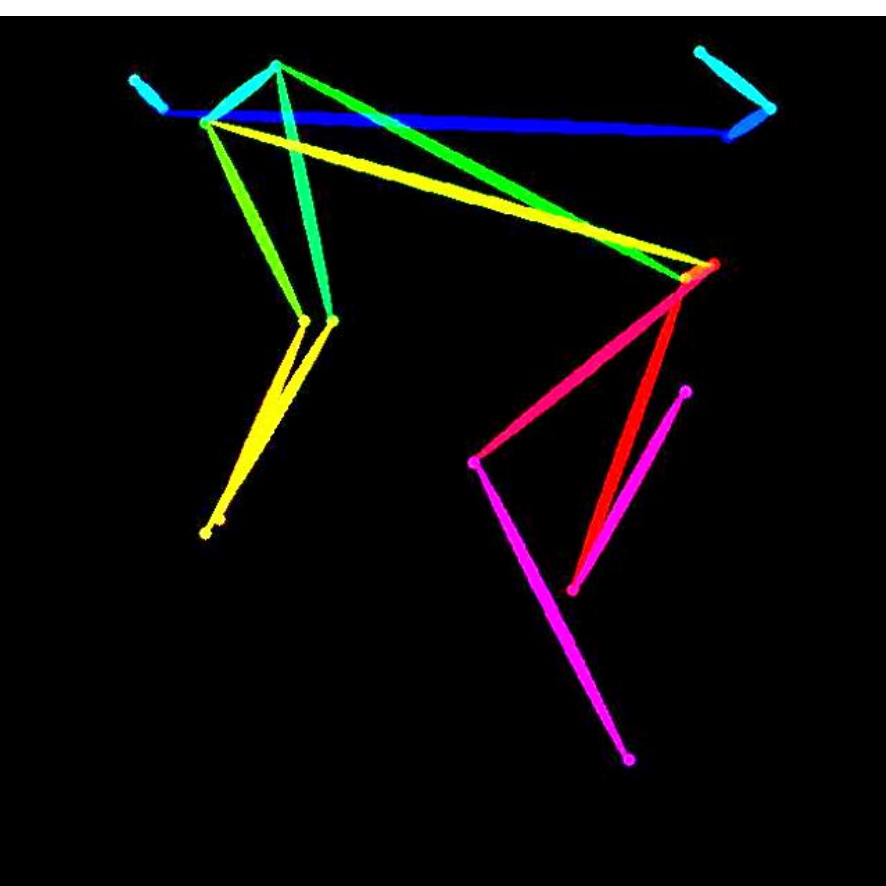}}\quad	
	\caption{Illustrations of the wrong skeletons generated by GANs-based method. Fig. \ref{fig:1(a)} shows a disproportionate skeleton, and Fig. \ref{fig:1(b)}shows a skeleton with misplaced key points, and Fig. \ref{fig:1(c)}shows a deformed and twisted skeleton.}
\label{fig:fig1}
\end{figure}

In recent years, many researchers have explored methods for generating 2D human pose skeletons from textual descriptions. \cite{zhang2021adversarial} trained a generative adversarial network capable of generating single-person poses from text. \cite{roy2022tips} proposed the DE-PASS, a dataset with detailed labelling of the facial details in poses, and trained a generative adversarial network for generating single-person poses conditioned on natural language on it, which designed a RefineNet to improve the generation of facial gestures. However, they share a common problem, i.e., the generated pose skeletons suffer from misplaced key points and disproportionate bone lengths. 
Fig. \ref{fig:fig1} shows several illustrations of erroneous skeletons generated by a GANs-based method(\cite{zhang2021adversarial}). Fig. \ref{fig:1(a)} demonstrates the disproportionate skeleton, where the two arms of the character are obviously not the same length. Fig. \ref{fig:1(b)} shows a skeleton with misplaced key points. The key point that should be on the feet or hidden mistakenly appeared near the shoulders, causing the calves to appear in the wrong position. Fig. \ref{fig:1(c)} demonstrates a deformed and twisted skeleton, in which the head key points are too widely dispersed, resulting in a deformed head, and the right knee is folded far beyond the limits of what humans can do, resulting in a partially deformed right leg.


To address the above challenges, we propose PoseDiffusion, a framework capable of generating 2D single-person pose skeletons from various texts. Inspired by the recent progress of the text-conditioned image generation(\cite{dhariwal2021diffusion,nichol2021improved,nichol2021glide,ramesh2022hierarchical}), we propose to introduce a denoising diffusion probabilistic model for tasks ranging from natural language to human pose skeleton generation. Unlike classical text-conditional image generation models that use 3- or 4-channel feature maps to represent an image as a whole, we propose to represent each key point in a 2D human pose skeleton as a separate feature channel, thus enabling individual prediction of the position of each key point. Similar to \cite{rombach2022high,dhariwal2021diffusion}, we propose to use cross-attention to combine input text and image noise. The above approach can significantly increase the diversity of the generated pose images. In addition, to avoid the model always generating deformed skeletons, PoseDiffusion's GUNet model treats the human skeleton as an undirected graph and introduces the key point and skeleton information into the generation process through a graph neural network. As a result, our model generates correct keypoint locations and appropriate skeleton lengths. Furthermore, in order to verify the effectiveness of introducing the skeleton information, we propose two fine-tuned Stable Diffusion-based variants. 

We perform extensive qualitative experiments and quantitative evaluations on popular benchmarks. First, we demonstrate significant improvements in text-driven 2D pose skeleton generation. Second, we show the generation capabilities of two fine-tuned variants. In addition, we discuss additional possibilities of PoseDiffusion for multiplayer pose generation.


In summary, our proposed PoseDiffusion has several desired properties that outperform the prior arts:
\begin{itemize}
\item \textit{1) Correct Skeletons.} Benefit from our designed denoising model GUNet, PoseDiffusion can use the connection information of the human skeleton as a reference during the denoising process, so as to generate human pose skeletons with the correct location of key points and appropriate length of the bones.
\item \textit{2) Diversity.} Benefit from our bold attempt at the conditional diffusion model,  decoupled representation of human posture skeleton, and cross-attention design of multimodal information, PoseDiffusion can achieve higher diversity in generating results.
\end{itemize}



\section{Related Works}
\subsection{Pose or Keypoint-guided Text-driven Image Generation}
With the advent of Stable Diffusion(\cite{rombach2022high})(SD), work on text-driven image generation has made tremendous progress. In order to avoid pose distortion of people or objects in the generated images, many works use pose skeletons to guide text-driven image generation. ControlNet(\cite{zhang2023adding}) incorporates additional inputs such as pose information into SD to enhance the architecture for controllable generation, allowing users to control image details during the generation process precisely. HumanSD(\cite{ju2023humansd}) is a model that improves on SD by introducing a specific human pose encode module to enhance the accuracy of human pose generation for fine-grained pose manipulation tasks. T2I-Adapter(\cite{mou2024t2i}) is an insertion module that combines textual prompts with additional visual guidance signals (e.g., depth images or pose images) to help the generative model better correspond complex textual descriptions to image content. GRPose(\cite{yin2024grpose}) combines graph convolutional neural networks(GNNs) and SD model to focus on generating multi-person pose characters, and processing the relationship of skeleton points through GNNs, which helps the SD model to better understand the pose image, and thus generates pictures with better pose adherence. All of the above work requires an additional picture of the pose skeleton. Current approaches mainly rely on extracting the pose skeleton from existing images using pose detection models such as DWPose(\cite{yang2023effective}) and OpenPose(\cite{cao2017realtime}), etc. This limits the diversity and tractability of the available pose skeletons.

To address this problem, we need to delve into its upstream work, i.e., synthesising high-quality human pose skeletons from diverse texts, thereby enriching the source of pose skeletons for text-driven image generation.

\subsection{Generating Human Pose Skeletons from Natural Language}
Previous work has done some research on generating human poses from natural language. \cite{reed2016learning} showed that higher-resolution images can be synthesised using a sparse set of key points. \cite{zhou2019text} change the pose of a person in a given image based on a textual description. However, the pedestrian dataset they used had simple poses and could not be applied to datasets with multifunctional and highly articulated poses, such as COCO. To enrich the generative results of the model, \cite{zhang2021adversarial} trained a generative adversarial network capable of generating complex poses from natural language descriptions on the COCO dataset. Although they achieved good results in terms of action complexity and versatility, they did not take into account the spatial relationships between different key points of human poses,  resulting in the generation of poses that are subject to problems such as deformities or skeletal disproportions. In addition, their simple treatment of text conditions results in generated poses that sometimes do not match the text well.

To solve these problems, we propose a new text-driven human pose skeleton generation pipeline based on denoising diffusion probabilistic model(DDPM)(\cite{ho2020denoising}). Where the denoising model is based on U-Net, and introduces a graph neural network to introduce the spatial information of the pose. In later sections, we will discuss the design of different variants of the U-Net-based diffusion model and the potential of our proposed pipeline in various application scenarios. Moreover, benefiting from the properties of the DDPM, our pipeline is able to generate more diverse samples.


\subsection{Text-conditioned U-Net}
U-Net was initially applied to solve the problem of medical image segmentation(\cite{ronneberger2015u,oktay2018attention,zhou2018unet++}), and more recent work combines U-Net with diffusion processes for image generation(\cite{rombach2022high,dhariwal2021diffusion,ju2023humansd,zhang2023adding}). These works achieve state-of-the-art performance on a variety of tasks such as unconditional image generation, text-driven image generation, and image-driven image generation, and inspired our work. Text-conditioned U-Net(\cite{rombach2022high}) is one of the pioneers in using U-Net as a denoising model. The basic idea of text-conditioned U-Net is to incorporate textual information into the feature interaction between the encoder and decoder layers of U-Net by introducing textual embedding and cross-attention mechanisms, thus effectively guiding the image content during the generation process. The down-sampling process of U-Net aims at gradually compressing the spatial dimension of the input feature map to extract higher-level, global semantic information, while the up-sampling process is used to gradually recover the spatial resolution to combine the global semantic information with the detailed features to generate the output image of the same size as the input. 

\section{Method}
We propose a diffusion model-based framework, PoseDiffusion, for generating diverse and skeletally structurally stable 2D human pose skeletons. We first give the problem definition in Section \ref{sec:3.1}, then we provide a general description of the proposed PoseDiffusion in Section \ref{sec:3.2}, followed by the diffusion model in Section \ref{sec:3.3} and the U-Net based denoising model, GUNet, in Section \ref{sec:3.4}.

\subsection{Task Definition}
\label{sec:3.1}
\begin{figure}[ht]
\begin{center}
\includegraphics[width=1\textwidth]{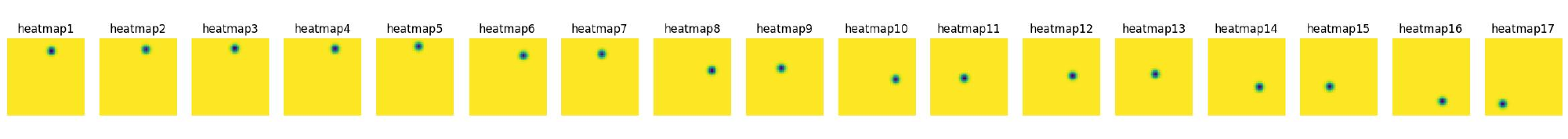}
\end{center}
\caption{Heatmap of a pose skeleton with 17 key points. }
\label{fig:fig2}
\end{figure}

The human pose skeleton $H$ contains a set of heatmaps $h_i$, where $i\in \left \{  1,2,...,K\right \}$ and $K$ denotes the number of key points in the skeleton. Each $h_i\in\mathbb{R}^{S\times S }$ corresponds to a heatmap for a keypoint, and $S$ is the size of these heatmaps. The heatmap is modelled with a Gaussian distribution centred on the coordinates of the keypoints to obtain a 2-dimensional image, where each pixel point represents the probability of occurrence of the key point, with darker colours indicating a higher probability of occurrence of the key point. We randomly selected a skeleton in the dataset containing 17 key points, and the heatmap visualization is shown in Fig. \ref{fig:fig2}. In text-driven human pose skeleton generation, the dataset consists of $\left ( H_i,text_i \right ) $ pairs, where $text_i$ is a natural language description of the human pose skeleton $H_i$. In the inference process, given a set of descriptions $\left \{text_i \right\} $ and a set of random noise samples with the same shape as $H$, the denoising model will generate a set of heatmaps from the random noise that match the given descriptions. In the following, we will use the abbreviation T2P(Text2Pose) to represent this task.


\subsection{Pipeline Overview}
\label{sec:3.2}
\begin{figure}[t]
\begin{center}
\includegraphics[width=1\textwidth]{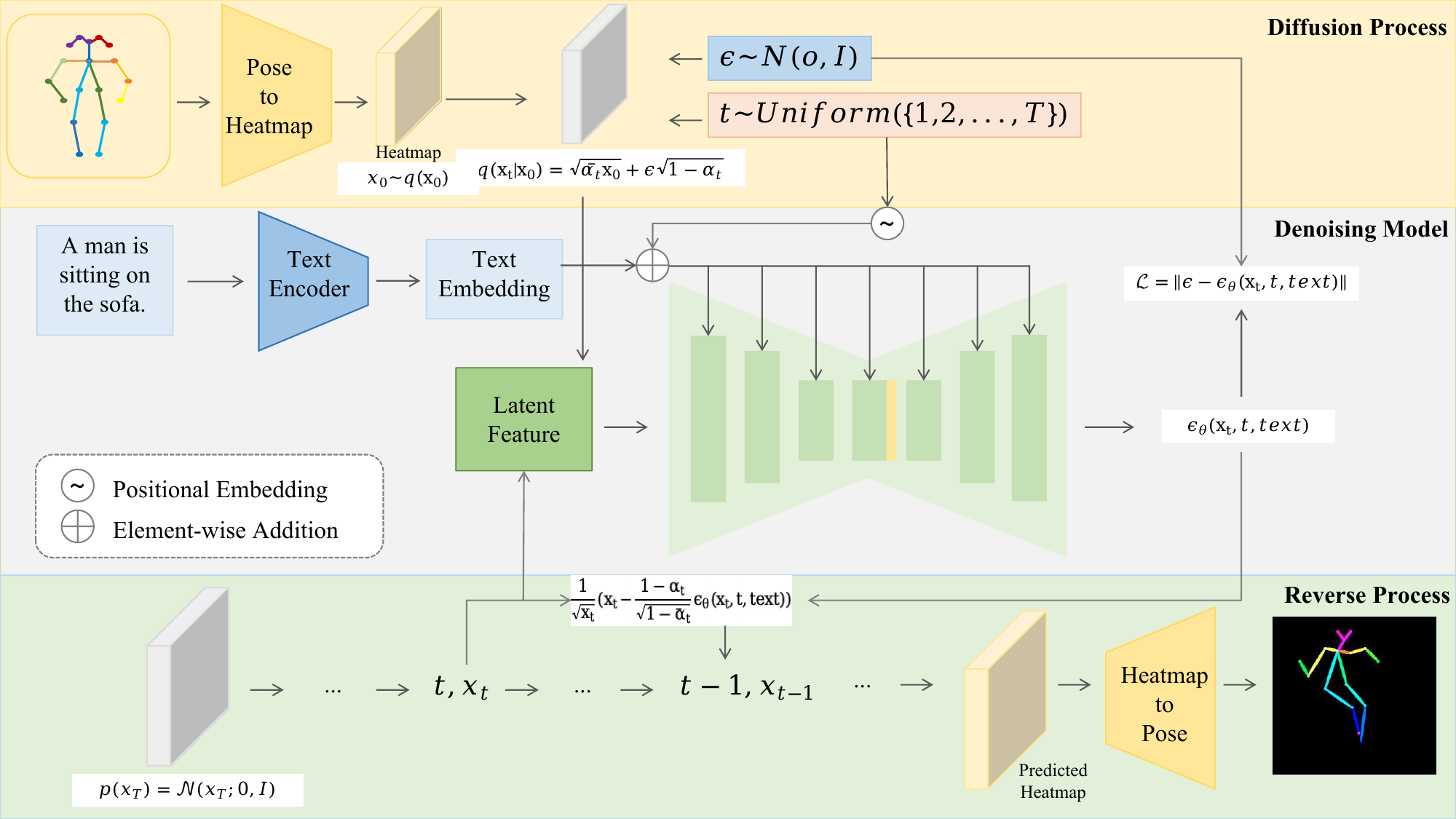}
\end{center}
\caption{Pipeline overview. In \textit{Diffusion Process}, we transform the pose skeleton into a set of heatmaps via the Pose2Heatmap module and then add noise to them on each timestep. The inputs to the \textit{Denoising Model} are noisy latent features from \textit{Diffusion Process} and text embedding with timestep embedding in the training process, and the output is the predicted noise on the input timestep. \textit{Reverse Process} samples the noise to obtain heatmaps that match the input text and transform them into the pose skeleton via the Heatmap2Pose module.}
\label{fig:fig3}
\end{figure}

Our proposed PoseDiffusion pipeline is shown in Fig. \ref{fig:fig3}. First, we construct a text-based pose skeleton generation pipeline using the denoising diffusion model(\cite{nichol2021improved})(DDPM). The basic of the denoising model is U-Net, based on which we insert a spatial module into U-Net to insert a spatial module into U-Net to introduce spatial information of the skeleton, such as key point locations and connectivity relationships, to obtain GUNet. To enable GUNet to handle textual conditions, we use a cross-attention mechanism to fuse cross-modal features. In addition to this, we change the input dimension of GUNet to match the skeleton features represented using a set of heatmaps. The above design significantly improves the pose skeleton generated by PoseDiffusion. We will describe each part of the pipeline in the following subsections. 

\subsection{Diffusion Model}
\label{sec:3.3}
The diffusion model is a generative model that progressively denoises Gaussian noise through a learnable probabilistic model. The forward process of diffusion is a Markov chain that starts from the initial data $x_0$ and gradually adds noise with variance $\beta_t$ to the data $x_{t-1}$ over $T$ timesteps to obtain a set of noise samples $x_t$ with distribution $q\left ( x_t\mid x_{t-1} \right )=N\left(x_t;\sqrt{1-\beta_t}x_{t-1},\beta_tI \right)$. The distribution of the whole forward noise addition process can be calculated by Eq. \ref{e1}:
\begin{equation}
\label{e1}
q\left ( x_{1:T}\mid x_0 \right )=q\left ( x_0\right )\prod_{t=1}^{T}q\left( x_t\mid x_{t-1}\right) 
\end{equation}
The inverse process is the reversal of the forward process, where we sample from $q\left ( x_t\mid x_{t-1} \right )$ and progressively reconstruct the true sample, which means estimating $q\left ( x_{t-1}\mid x_t \right )$ at the moment $t = T$. Estimating the previous state from the current state requires knowledge of all the previous gradients. Thus, it is necessary to train a neural network model to estimate $p_{\theta}\left ( x_{t-1}\mid x_t \right )$ based on the learned weights $\theta $ and the current state at time $t$. This trajectory can be performed by Eq. \ref{e2}:
\begin{equation}
\begin{split}
\label{e2}
p_{\theta}\left ( x_{t-1}\mid x_t \right )=N\left(x_{t-1};\mu _\theta \left( x_t, t\right), {\textstyle \sum_{\theta}^{}} \left( x_t, t\right)\right)\\
p_{\theta}\left( x_{0:T}\right)=p\left( x_T\right)\prod_{t=1}^{T}p_{\theta}\left( x_{t-1}\mid x_t\right) 
\end{split}
\end{equation}
To provide a simplified representation of the diffusion process, \cite{ho2020denoising} formulated it as Eq. \ref{e3}:
\begin{equation}
\label{e3}
q\left(x_t\mid x_0 \right)=\sqrt{\bar{\alpha_t } }x_0+  \sqrt{1-\bar{\alpha _t} } ,\epsilon \in N\left ( 0,I \right ) 
\end{equation}
where $\alpha_t=1-\beta_t$ and $\bar{\alpha _t}= {\textstyle \prod_{s=0}^{t}}\alpha   _s$. Thus we can simply sample the noise samples and generate $x_t$ directly from this formula. Here, we follow the approach used in GLIDE(\cite{nichol2021glide}) and predict the noise term $\epsilon$. The expression for the neural network prediction during the sampling process can be simplified to $\epsilon_{\theta}\left( x_t,t,text\right)$. We optimize the denoising model using the loss function as shown in Eq. \ref{e4}:
\begin{equation}
\label{e4}
\mathcal{L}=E_{t\in\left[1,T\right],x_0\sim q\left(x_0\right),\epsilon \sim\mathcal{N}\left(0,I\right)}\left[\left \| \epsilon -\epsilon _{\theta}\left(x_t,t,text\right) \right \| \right]
\end{equation}
To generate samples from a given textual description, we perform noise reduction on the sequence from $p\left(x_T\right)=N\left(x_T;0,I\right)$. From Equation. \ref{e2}, we know that we need to estimate $\mu_{\theta}\left(x_t,t,text\right)$ and $\sum_{\theta}\left(x_t,t,text\right)$. To simplify this, we set $\sum_{\theta}\left(x_t,t,text\right)$ to a constant $\beta_t$, and $\mu_{\theta}\left(x_t,t,text\right)$ is approximated as Eq. \ref{e5}:
\begin{equation}
\label{e5}
\mu_{\theta}\left(x_t,t,text\right)=\frac{1}{\sqrt{x_t} }\left(x_t-\frac{1-\alpha _t}{\sqrt{1-\bar{\alpha _t} } }\epsilon _{\theta}\left(x_t,t,text\right) \right) 
\end{equation}

\subsection{GUNet and Components}
\label{sec:3.4}
In the previous section, we introduced the conditional diffusion model, highlighting the critical role of the denoising network $\epsilon_{\theta}\left( x_t,t,text\right)$ in the denoising process. In this section, we present the denoising network of PoseDiffusion, GUNet.

\begin{figure}[t]
\begin{center}
\includegraphics[width=0.8\textwidth]{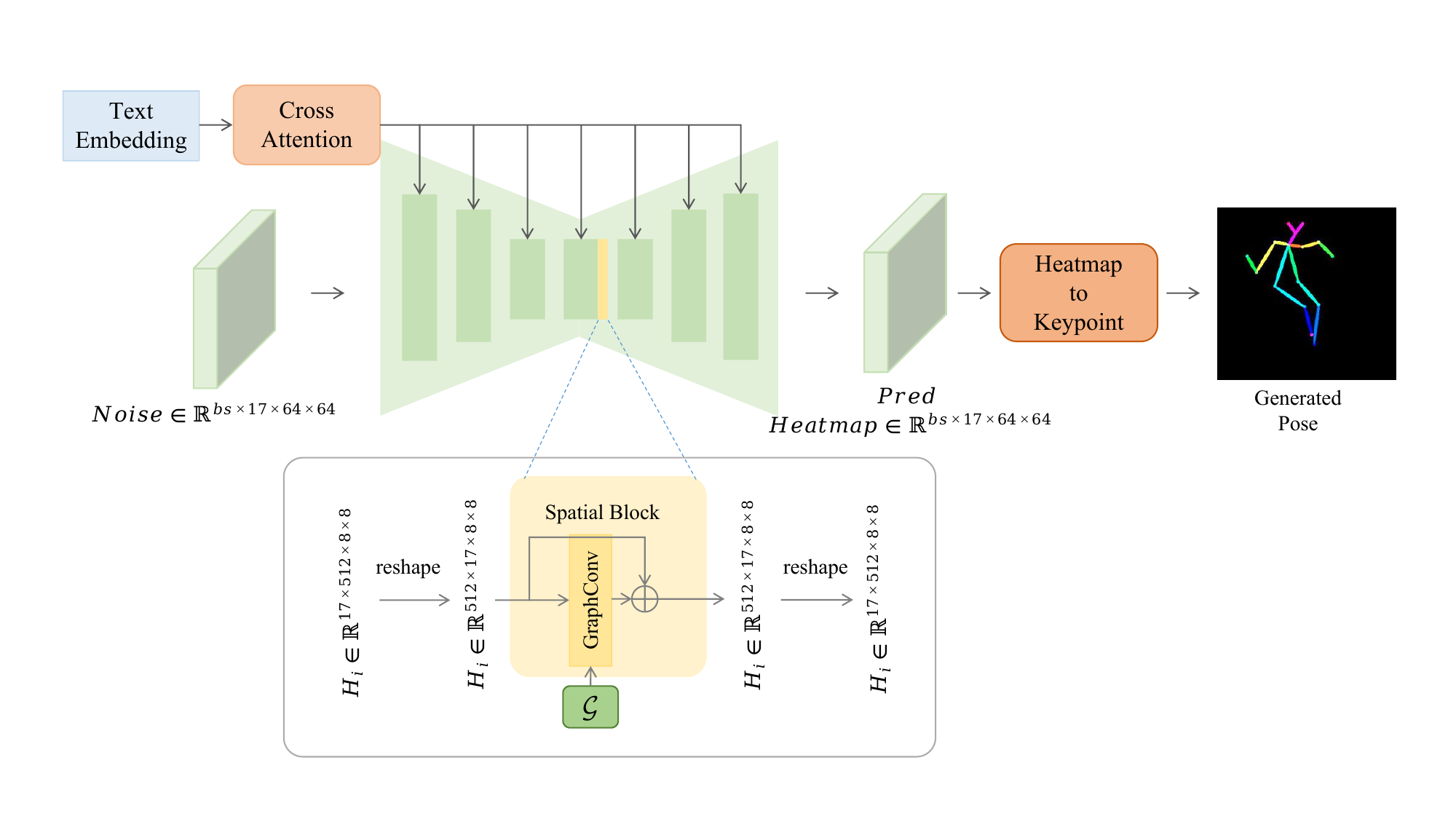}
\end{center}
\caption{GUNet. GUNet is a U-Net like structure consisting of three downsampling blocks, three upsampling blocks, an middle block and a spatial block. The sampling blocks and the middle block consist of CNN layers, a self-attention layer, and a cross-attention layer. The middle block is followed by a spatial block containing a graph convolutional neural network layer and a skip connection.}
\label{fig:fig4}
\end{figure}

Previous work(\cite{ho2020denoising,dhariwal2021diffusion,nichol2021glide,nichol2021improved}) using a U-Net like structure as a denoising model. Since our skeleton generation task requires the introduction of spatial information about the skeleton during the denoising process, it makes CNNs, a structure for processing image data, unable to explicitly model the pose skeleton. In order to enable diffusion models to process data with explicit spatial architectures, \cite{wen2023diffstg} proposed the use of graph neural networks as a building block for U-Net like structures. However, the representation of the pose skeleton, heatmap, as a kind of 2D image data, graph convolutional neural networks, a structure dedicated to processing structured data, are unable to perform detailed feature extraction on it. Therefore, we propose a U-Net like structure GUNet that combines the advantages of both CNN and GNN, as shown in Fig. \ref{fig:fig4}. Similar to the denoising model of text-driven image generation, our proposed model also includes a text encoder, a pose encoder and a pose decoder.

\textbf{Text Encoder.} In this paper, we use BERT(\cite{devlin2018bert}) to obtain the embedding of the text. Specifically, the sentence $T$ describing a pose skeleton is firstly be spilt by the tokenizer of BERT to get $T_{token}=\left \{  t_1,t_2,.t_L..,\right \}$. Then, $T_{token}$ is input to the embedding layer of BERT for encoding, using the embedding of CLS token as the sentence representation, i.e., $T\in\mathbb{R}^{1\times 768 }$.

\textbf{Pose Encoder and Pose Decoder.} Pose Encoder and Decoder are the Pose2Heatmap and Heatmap2Pose modules in Fig. \ref{fig:fig3}. These are two modules that require no training and enable the transformation between a pose skeleton and a set of heatmaps corresponding to it. Specifically, the COCO dataset comes with keypoint coordinates for each pose skeleton of the class. We generate a heatmap of size $S\times S$ for each keypoints of the pose skeleton, and the value of each pixel point of the heatmap represents the probability that the keypoint occurs here. Assuming that the coordinate of the key point $K_i$ is $\left( \mu_x, \mu_y\right)$, then, the pixel value at coordinate $\left(x,y\right)$ is computed by Eq. \ref{e6}
\begin{equation}
\label{e6}
heatmap\left(x,y\right) = e^{-\frac{{\left(x-\mu_x\right)}^2+{\left(y-\mu_y\right)}^2}{2\sigma^2} }
\end{equation}
Eventually, each pose skeleton is represented as a set of heatmaps by PoseEncoder(Pose2Heatmap), $H\in\mathbb{R}^{K\times S\times S}$, where $K$ denotes the number of keypoints of a skeleton, and $S$ denotes the size of the heatmap. PoseDecoder(Heatmap2Pose) directly obtains the coordinates of the largest pixel value on the heatmap as the keypoint coordinate. Finally, the key points are connected and coloured using OpenPose's(\cite{cao2017realtime}) skeleton connection and color rules to get the pose skeleton.

\textbf{GUNet.}  In GUNet, we define the human pose as a graph structure, notated as $\mathcal{G}=\left(V,E,A\right)$, where $V$ represents the set of node features, $E$ represents the set of edges, and $A$ is the adjacency matrix defining the relationships between nodes, defined as Eq. \ref{e7}:
\begin{equation}
\label{e7}
A_{ij}=\begin{cases}
  & 1,\text{ if there is an edge between node $i$ and node $j$}   \\
  & 0,\text{ others }
\end{cases}
\end{equation}
Although the graph structure is fixed for different human pose skeletons, as the pose changes, the associated heatmap sets will be different and the features of the node set $V$ will be updated.

As shown in Fig. \ref{fig:fig4}, GUNet is a U-Net like structure consisting of three down-sampling blocks, three up-sampling blocks, one middle block and one spatial block. Consistent with previous work(\cite{nichol2021improved}), both the downsampling, upsampling and middle blocks consist of CNN layers, and each block contains a self-attention layer and a cross-attention layer for maintaining cross-modal semantic consistency between the textual conditions and the images. We insert a spatial block containing a graph convolutional layer(GCN) and a skip connection after the middle block. The reason is that the image features include a lot of detailed information at the original resolution. The main network structure in the spatial block, GCN, is good at dealing with graph data structures, and its nodes tend to be low-dimensional feature representations. 

When the latent embedding $N$ reaches the spatial block of GUNet, we rearrange the dimensions of $N \in \mathbb{R}^{bs \times K_{MID} \times S_{MID} \times S_{MID}}$ to $N \in \mathbb{R}^{K_{MID} \times bs \times S_{MID} \times S_{MID}}$ to facilitate smooth graph convolution computation. The rearranged latent embedding, which serves as the feature for the node set in the graph structure, is then fed into the graph convolution layer. The graph convolution can be formulated as shown in Eq.\ref{e8}.
\begin{equation}
\label{e8}
\Gamma _{\mathcal{G}}\left(\bar{N} \right)=\sigma \left(\Phi \left(A_{gcn},N\right)W\right)
\end{equation}
where $W\in\mathbb{R}^{bs\times bs}$ denotes the trainable parameters, $\sigma$ is the activation function, and $\Phi \left(\cdot \right)$ is the rule aggregation function that determines how neighbouring features are aggregated into the target node. In this function we directly use the form in the most popular vanilla GCN, defining a symmetric normalised summation function as $\Phi \left(A_{gcn},N \right)=A_{gcn}N$, where $A_{gcn}=D^{-\frac{1}{2} }\left(A+I\right)D^{-\frac{1}{2} }\in\mathbb{R}^{V\times V}$ is the normalised adjacency matrix of the graph $\mathcal{G}$, $I$ is the unit array, and $D$ is the diagonal matrix, where $D=\left(A+I\right)$. The output of the GraphConv layer is then summed with its input to achieve the skip connection, ensuring the stability of the features. Finally, the positions of $bs$ and $K_{MID}$ in $N\in\mathbb{R}^{K_{MID}\times bs\times S_{MID}\times S_{MID}}$ are exchanged again to produce the output of the spatial module, which is then upsampled by GUNet to generate the predicted heatmap.

\section{Experiments}
In this section, we discuss the experimental design and its results. We selected two baseline models based on GAN(\cite{zhang2021adversarial}): WGAN-LP Regression (for regression prediction) and WGAN-LP (for heatmap prediction). In addition, we included several baseline models proposed in this paper, including SD1.5-T2Pose(\cite{rombach2022high}) with full fine-tuning, PoseAdapter adapted from IPAdapter(\cite{ye2023ip}), and UNet-T2H, i.e., GUNet without GCNs. We provide a comprehensive comparison of these models with the GUNet proposed in this paper. The dataset, descriptions of the baseline models, and the qualitative and quantitative results are presented separately below.

\subsection{Dataset}
We use the COCO (\cite{lin2014microsoft}) to train and evaluate our models. This dataset contains over 100,000 annotated images of everyday scenes, each accompanied by five natural language descriptions. Among these, about 16,000 images feature a single person, with each person annotated by the coordinates of 17 keypoints. To ensure data quality, we filtered approximately 12,000 single-person images, selecting only those with at least 8 visible keypoints. For each image, one of the five descriptions was randomly selected as the image's label, forming image-text pairs. These pairs were then divided into training and validation sets with a 4:1 ratio. The dataset format for our baseline models differs slightly and will be detailed in Section \ref{sec:4.2}.

\subsection{Baseline Models}
\label{sec:4.2}
Our baseline models come from three main sources. First, we fine-tuned Stable Diffusion in two ways: SD1.5-T2P and PoseAdapter, which introduces a pose coding layer. Second, we modified GUNet by removing the graph convolutional neural network layer, resulting in a model we denote as UNet-T2H, where the human posture skeleton is represented as a set of heatmaps. Third, we included two models based on WGAN, focusing on different methods of human posture skeleton representation: WGAN-LP for heatmap prediction and WGAN-LP R for coordinate regression prediction. Next, we will describe the structure and experimental setup of each baseline model in detail.

\subsubsection{WGAN-LP \& WGAN-LP R}
WGAN-LP and WGAN-LP R are derived from the related work(\cite{zhang2021adversarial}). We follow its dataset and training settings to train the two models, WGAN-LP and WGAN-LP R. The training environment is consistent with the aforementioned models. 

\subsubsection{SD1.5-T2P \& PoseAdapter}
SD1.5-T2P and PoseAdapter are two text-driven pose generation models obtained by fine-tuning Stable Diffusion 1.5. We saved the skeleton as RGB images with the size of $256\times 256 $to fine-tune SD1.5 at full volume to obtain SD1.5-T2P. Inspired by IPAdapter, PoseAdapter adds a pose coding module to SD1.5-T2P to introduce the skeletal connection of human posture. During the training of SD1.5-T2P, we updated the parameters of UNet in Stable Diffusion, while freezing the parameters of VAE, CLIP and other components. In PoseAdapter, we have designed a pose encoding block consisting of a linear layer and a normalisation layer. During the training process, the pose embedding block encodes the skeleton features to obtain the pose embedding, which is then connected to the text embedding and used together as inputs to the conditional bootstrap denoising model. The fine-tuning process is similar to SD1.5-T2P. We selected approximately 5,000 images with clearly visible skeleton connections, which were then partitioned into training and testing sets in a 4:1 ratio.

\subsubsection{UNet-T2H \& GUNet}
UNet-T2H is the base version of the GUNet model without the graph convolutional layer, and we train conditional UNet from scratch using the COCO dataset. We use the Pose2Heatmap module mentioned in section \ref{sec:3.2} to encode the coordinate information of the human postural skeleton in the COCO dataset into heatmaps of size $17\times 64 \times 64$, which are paired with the corresponding textual descriptions, totalling about 12,000 pairs. The datasets are used in a 4:1 ratio for training and testing respectively.

The experimental results of the baseline model will be discussed in Sections \ref{sec:4.3} and \ref{sec:4.4}.

\subsection{Qualitative results}
\label{sec:4.3}

We conducted a series of generative experiments using the above models. The design and results of these experiments are discussed below.

\begin{figure}[htbp]
    \centering
    \includegraphics[width=0.8\textwidth]{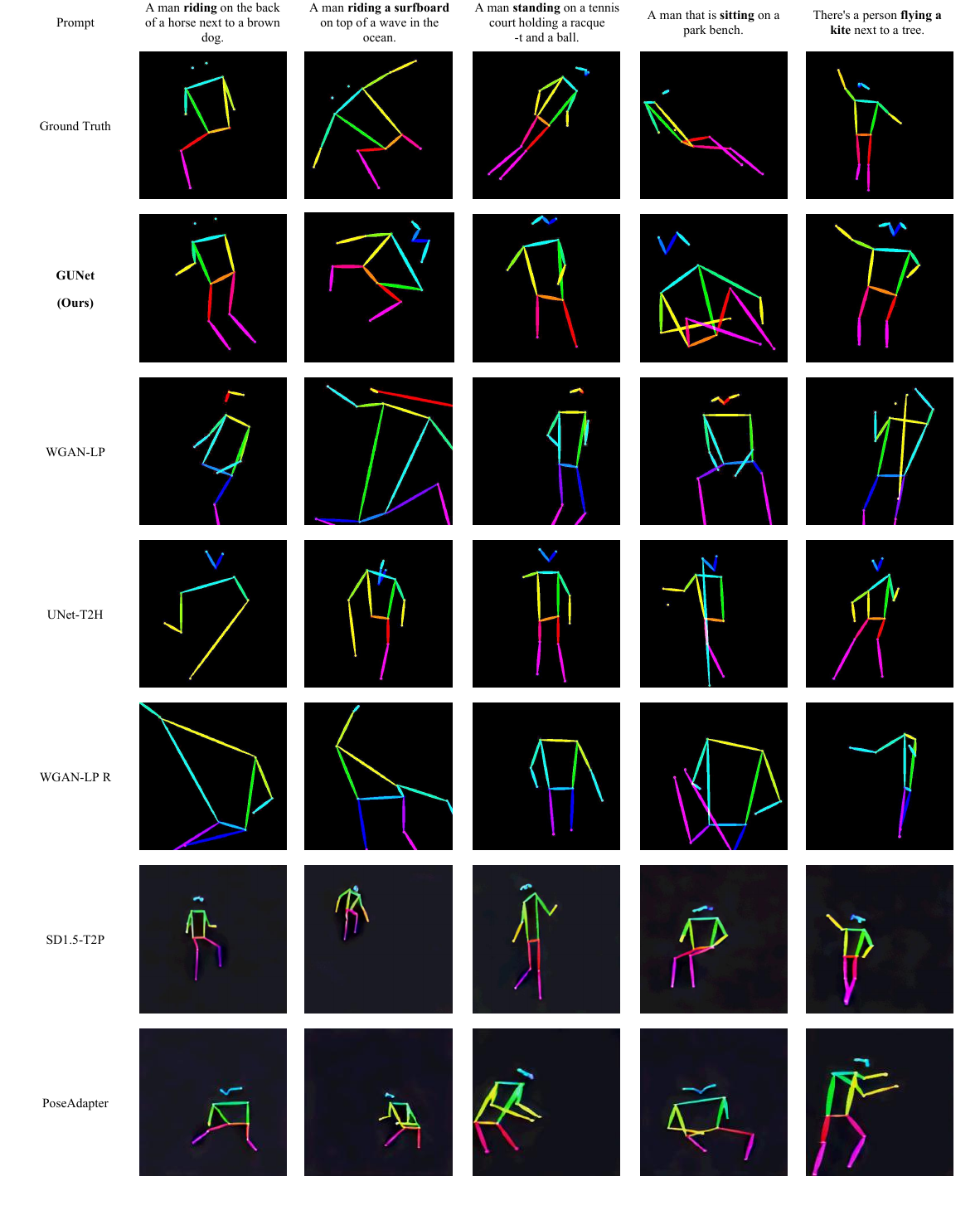}
    \caption{Some qualitative poses generated by the model, using the ground truth pose in the first line as a reference.}
    \label{fig:fig5}
\end{figure}

\begin{figure}[htbp]
	\centering
	\subfloat[]{\label{fig:6(a)}\includegraphics[width=0.68\textwidth]{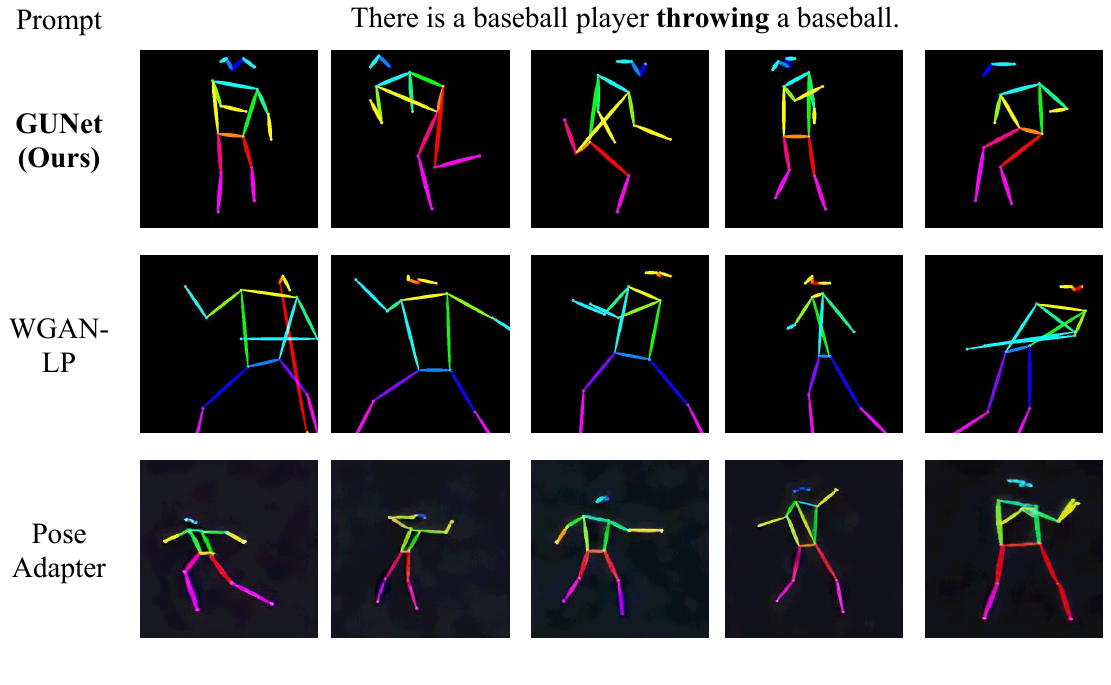}}\quad
	\subfloat[]{\label{fig:6(b)}\includegraphics[width=0.68\textwidth]{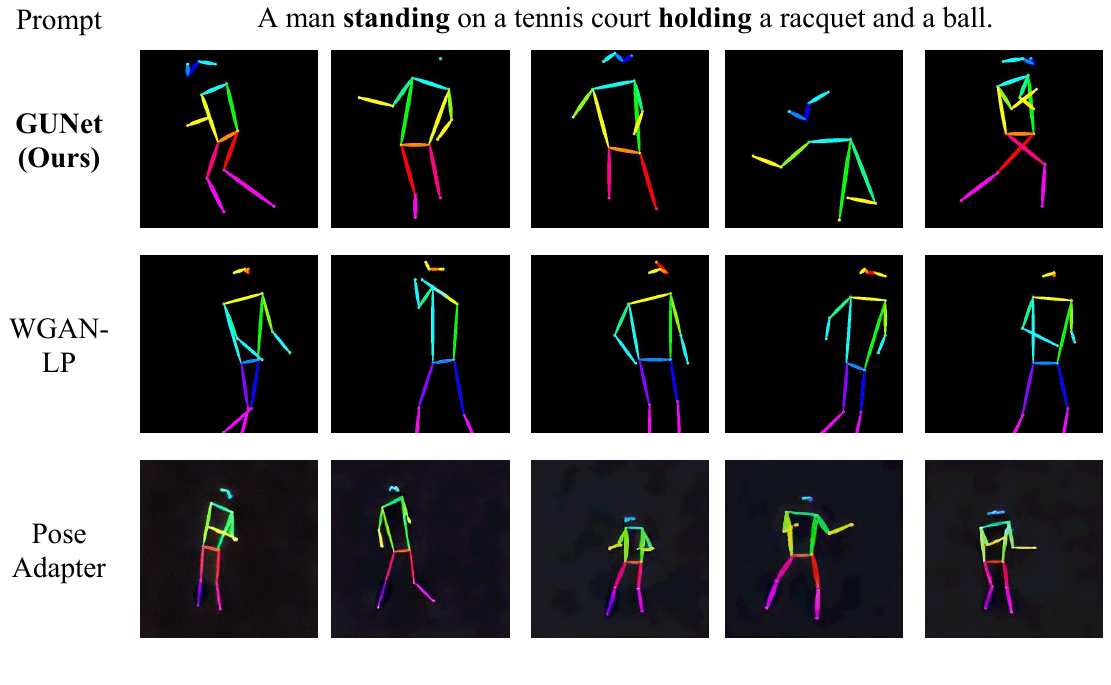}}\quad	
	\caption{Comparison of the diversity of poses generated by different models.}
\label{fig:fig6}
\end{figure}
Firstly, we take five text descriptions randomly from the validation set, including common daily actions such as ride, stand, sit, fly a kite, etc., and generate pose skeletons for each using the model described in Sec.\ref{sec:4.2}. We then compared these generated skeletons with ground truth.
The results in Fig.\ref{fig:fig5} show that the human pose skeleton generated by our GUNet (second row of Fig. \ref{fig:fig5}) conforms to the given textual description. And, our results show fewer misplaced keypoints and disproportionate skeletons than other baseline models. 

To demonstrate the effectiveness of introducing the spatial relationship of the human skeleton, we compare the results of GUNet (the second row of Fig.\ref{fig:fig5}) with those of UNet-T2H (the fourth row of Fig. \ref{fig:fig5}). The UNet-T2H results exhibit clear proportion issues, such as the arms being noticeably longer than normal in the first and second poses of the fourth row, whereas GUNet’s results do not have this issue. 

To verify the superior performance of the diffusion model compared to GAN, we compare the results of GUNet (the second row of Fig. \ref{fig:fig5}) with those of WGAN-LP (the third row of Fig. \ref{fig:fig5}). It can be seen that the GUNet results are more uniform and coordinated in terms of the distribution of the torso and limbs, such as those generated in the second column, while the poses generated by WGAN-LP are more localised, which is not conducive for subsequent models such as ControlNet
to analyse the semantics of the poses.

To evaluate the diversity of human pose skeletons generated by different models, we selected two different scenarios of action descriptions and used each model to generate five human pose skeletons. Since it was demonstrated in the previous paragraph that the model without skeleton connection information significantly underperforms the model with skeleton connection informationn, we simplify the experiments by using only GUNet, WGAN-LP and PoseAdapter for generation.
The results in Fig. \ref{fig:fig6} show that the pose skeletons generated by WGAN-LP and PoseAdapter exhibit high similarity under the same textual description, i.e., not much of the change in the main skeleton, whereas the pose skeletons generated by GUNet displays greater diversity and variability.

\begin{figure}[htbp]
\begin{center}
\includegraphics[width=0.8\textwidth]{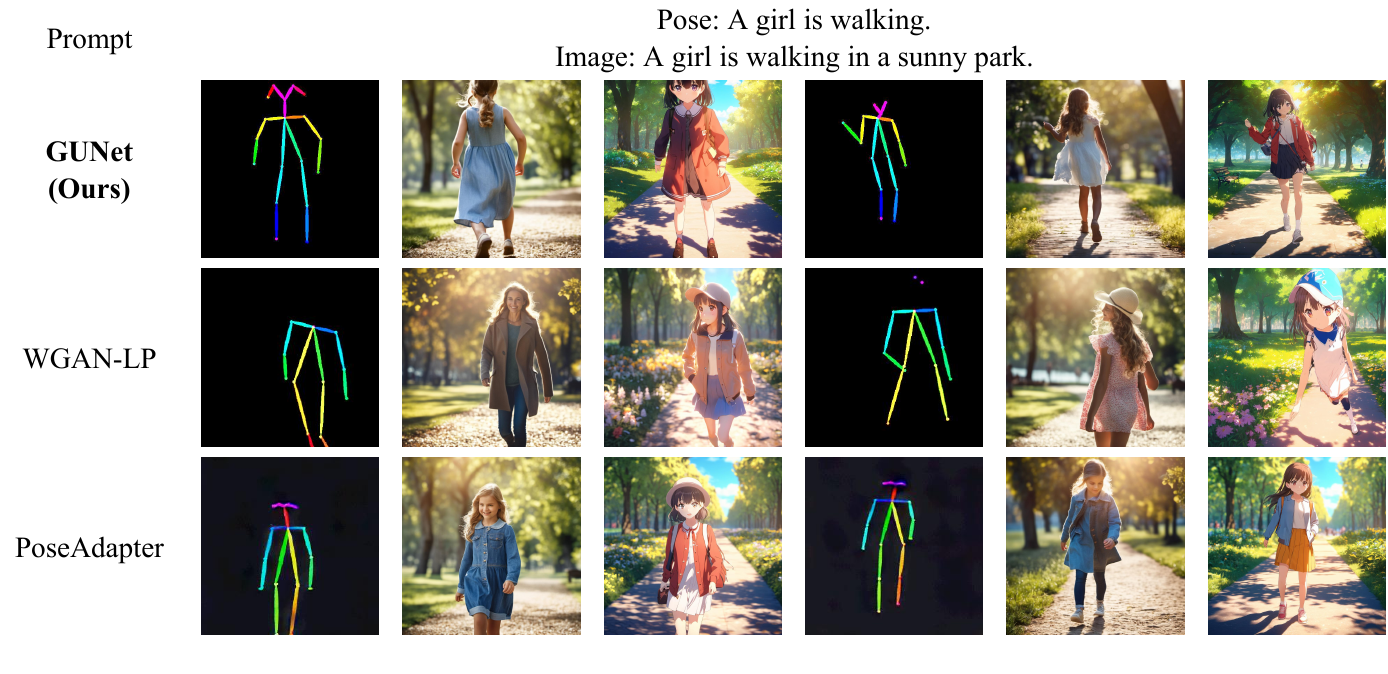}
\end{center}
\caption{Comparison of the aesthetics of different model-generated poses combined with Controlnet to generate real images.}
\label{fig:fig7}
\end{figure}
Finally, in order to verify the effect of the quality of the human pose skeleton on the quality of ControlNet-generated images, we generated human pose skeletons for different models using the same prompts, and then combined them with ControlNet to generate images via stable diffusion.ChatGPT rewrites and enhances the cues used for stable diffusion. As can be seen in Fig.\ref{fig:fig7}, in the images generated from the pose skeletons generated by WGAN-LP, the poses of the characters often do not match the pose skeletons. It is necessary to point out here that the images generated with reference to the pose map do not exactly follow the structure of the pose map, as mentioned in \cite{zhang2023adding}. However, the pose skeleton generated by WGAN-LP in combination with ControlNet is prone to errors such as multiple arms in the generated images, which reflects that the pose skeleton generated by WGAN-LP deviates from the real pose distribution, does not conform to the scientific human skeletal structure, and is not coordinated enough as a whole. On the contrary, we introduced the human skeleton information in GUNet and PoseAdapter, which makes the pose of the character in the picture generated by ControlNet match well with the skeleton generated by GUNet and PoseAdapter.

\subsection{Quantitative results}
\label{sec:4.4}

The previous section qualitatively analysed the performance of different models by observing the results generated by the models. In this section, we will design a series of experiments for quantitative evaluation to further validate our conclusions.
\begin{table}[htbp]
	\caption{Quantitative Result 1}
	\centering
	\begin{tabular}{lll}
		\toprule
            Methods & MSE & Var \\
            \midrule
		\textbf{GUNet(Ours)} & \textbf{262.2} & \textbf{244.5} \\
            UNet-T2H & 278.4 & 233.4  \\
            WGAN-LP & 286.9 & 172.2  \\
            WGAN-LP R & 290.8 & 131.2 \\
		\bottomrule
	\end{tabular}
	\label{tab:table1}
\end{table}

To validate the accuracy and diversity of model-generated poses, this paper uses MSE and variance as quantitative assessment metrics. Since models based on Stable Diffusion fine-tuning generate pose images directly without first generating intermediate state keypoints, we calculated MSE and variance for four models: GUNet, UNet-T2H, WGAN-LP, and WGAN-LP R. First, we generated 10 pairs of human posture skeletons for the natural language descriptions in the validation set with these models and then compared the coordinates of each keypoint of the generated posture skeletons with the corresponding ground truth to calculate MSE, taking the average value as the accuracy score. Second, we calculated the variance between the coordinates of these 10 pairs of posture skeletons and took the average value on the validation set as the diversity metric. The experimental results are shown in the first and second columns of Table. \ref{tab:table1}, where GUNet excels in both MSE and variance metrics, achieving the smallest MSE and the largest variance, representing the highest stability and diversity achieved.

To compare the aesthetic scores of different model-generated pose skeletons and ControlNet-generated images combined in Stable Diffusion, we employ Hpsv2(\cite{wu2023human}) as a quantitative evaluation tool and invite users to make preference choices through a blind selection evaluation. Hpsv2 is a state-of-the-art benchmarking framework for evaluating text-to-image generative models based on the large-scale human preference dataset HPDv2, designed to accurately predict human preferences for generating images. As shown in Table. \ref{tab:table2}, we performed Hpsv2 on GUNet, WGAN-LP, and PoseAdapter with scores of 0.2535, 0.2446, and 0.2561, respectively. These results are consistent with the intuition from the qualitative evaluation in Sec.\ref{sec:4.3}, where the images generated by PoseAdapter combined with ControlNet have the highest aesthetic scores, followed by those generated by GUNet, with WGAN-LP having the lowest. This further reflects that with the introduction of human pose skeleton information, the pose skeleton generated by the models can be more in line with the scientific human body structure, and thus perform better in the subsequent process of guiding the generation of real person images. WGAN-LP, on the other hand, does not introduce the human pose information, thus leading to poor subsequent performance.

\begin{table}[htbp]
	\caption{Quantitative Result 2}
	\centering
	\begin{tabular}{llll}
		\toprule
            Methods & Hpsv2 & User Preference(Pose) & User Preference(Image) \\
            \midrule
		\textbf{GUNet(Ours)} & 0.2535 & 43\% & 31\% \\
            WGAN-LP & 0.2446 & 7\% & 22\% \\
            PoseAdapter & \textbf{0.2561} & \textbf{50\%} & \textbf{47\%} \\
		\bottomrule
	\end{tabular}
	\label{tab:table2}
\end{table}

In the blind selection evaluation section, we designed two experiments for blind selection evaluation of aesthetic preferences for posture skeleton and ControlNet-generated pictures. 

In the pose skeleton experiment, we use three models to generated pose skeletons for a batch of pose description texts, and ask the users to choose the one that they think is in line with the scientific distribution of human bones, as well as the most aesthetically pleasing skeleton. From the results in the second column of Table. \ref{tab:table2}, the percentages of user preference for pose skeleton are 43\%, 7\% and 50\%, respectively. The percentages of PoseAdapter and GUNet are much higher than that of WGAN-LP, indicating that our model generates a more scientific and aesthetically pleasing pose skeleton from the user's point of view. 


In the ControlNet experiments, we use the pose skeleton generated by the three models as a reference picture for ControlNet to generate a batch of real pictures, and ask the user to choose a picture that he thinks the pose of the person in the real picture best matches the pose in the reference picture and is most aesthetically pleasing. From the results in the third column of Table. \ref{tab:table2}, the percentages of user preference for pose skeleton are 31\%, 22\% and 47\%, respectively. Although GUNet is rated lower than PoseAdapter, overall, our Diffusion-based scheme far outperforms WGAN-LP in terms of the percentage of preferences used, which suggests that a scientifically aesthetically pleasing skeleton is important for guiding the effectiveness of controlled image generation. In addition, by analysing the difference between GUNet and PoseAdapter in terms of user preferences, we conclude that PoseAdapter performs better in terms of compatibility with ControlNet and SD since it is a Stable Diffusion fine-tuned based model. Since GUNet is decoupled generation for each key point, it has a wider application prospect compared to PoseAdapter, e.g., multiplayer pose generation.

\section{Conclusions}
In this paper, we present PoseDiffusion, a text-driven human pose skeleton generation architecture based on a diffusion model, comprising GUNet and its two Stable Diffusion-based variants. GUNet is an innovative approach to human pose generation based on diffusion models and graph neural networks. We first describe the rationale behind choosing the UNet architecture as the foundation for the diffusion model and explain the effectiveness of representing human skeletons using heatmaps. To further enhance the model's performance, we introduce a graph convolutional layer into the UNet, creating GUNet, which better captures the spatial correlations and pose information in skeleton generation.

In the experimental section, we detail the design and experimental setup of the two variants based on Stable Diffusion. We conducted extensive experiments on the COCO dataset to validate the effectiveness of the proposed model. The results show that PoseDiffusion (containing GUNet and two SD variants) outperforms existing GAN-based baseline models in terms of accuracy and diversity of pose skeleton generation. In addition, when integrated with ControlNet to generate realistic images, PoseDiffusion achieved higher aesthetic scores compared to the GAN-based model. In the final blind user preference test, PoseDiffusion obtained higher user preference percentages for both the human pose skeleton and the realistic image. This indirectly reflects that high-quality skeleton generation can improve the aesthetic quality of subsequent image generation.

While one of the SD-based variants, PoseAdapter , performs well in blind user evaluation experiments, on the one hand, it was able to better compatibilise with ControlNet, partly because it was directly fine-tuned from SD. On the other hand, generation due to the fact that it treats the pose skeleton as a picture for overall generation leads to an overdependence on the quality of the training data and the inability to further optimise it by decoupling the keypoints in the event of generating incorrect connections. Based on this, although PoseAdapter can perform well in ControlNet, there is extremely limited room for future expansion and optimisation. On the contrary, GUNet benefits from the decoupling of key points, which makes it more advantageous in the tasks of multi-person pose generation and enhancing site-specific pose control.

\section{Future Work}
\label{sec:6}
The decoupling of keypoint representations in GUNet allows it to outperform PoseAdapter in multi-person pose skeleton generation tasks. Future work should further explore the potential of GUNet in multi-person pose generation and tasks requiring precise control over specific body parts. In current multi-person pose generation methods, there is no way to distinguish which keypoints belong to which individual, leading to unsatisfactory results in generating interactions such as hand-holding or hugging. Extending GUNet to multi-person pose skeleton generation would enable precise control over the keypoints of different individuals.

For tasks requiring precise control of specific body parts in human poses, we could map textual descriptions of different body parts in the prompt to corresponding pose regions by adjusting the prompt and incorporating masks. This approach would allow for more accurate manipulation of body part positioning based on text descriptions.


\bibliography{references}
\bibliographystyle{unsrtnat}


\end{document}